\documentclass[arxiv]{sparreport}
\usepackage{amsmath}
\usepackage{amssymb}
\usepackage{natbib}
\usepackage{tikz}
\usepackage{pgfplots}
\usepackage{pgfplotstable}
\usepgfplotslibrary{fillbetween}
\usepackage{xurl}   
\usepackage{hyperref}
\usepackage{booktabs}
\hypersetup{
    bookmarksnumbered,      
    bookmarksopen=true,     
    bookmarksopenlevel=3,   
    colorlinks,             
    allcolors=black,        
    citecolor=[HTML]007f00, 
    urlcolor=[HTML]00007f,  
    linkcolor=[HTML]b70000, 
    pdftitle={Infra-Bayesian Reinforcement Learning Agents Outperform Classical RL For Worst-Case Robustness},
}

\title{Infra-Bayesian Reinforcement Learning Agents Outperform Classical RL For Worst-Case Robustness}

\author{
    \textbf{Manish Aryal}\footnotemark[1] \\
    \texttt{aryalm@purdue.edu} \\
    Purdue University
    \and
    \textbf{Faiyaz Azam}\footnotemark[1] \\
    \texttt{fazam@andrew.cmu.edu} \\
    Carnegie Mellon University
    \and
    \textbf{Agnivo Banerjee}\footnotemark[1] \\
    \texttt{agnivo.stem.iiserk@gmail.com} \\
    WorldQuant University
    \and
    \textbf{Syed Mahir Ahamed}\footnotemark[1] \\
    \texttt{mahirahamed17@gmail.com} \\
    Union College
    \and
    \textbf{Sai Sidhanth Manoharan Jayanthi}\footnotemark[1] \\
    \texttt{sidmanoharan\_2512@berkeley.edu} \\
    UC Berkeley
    \and
    \textbf{Allegra Laro}\footnotemark[1] \\
    \texttt{allegralaro@gmail.com} \\
    Independent
    \and
    \textbf{Clément Legentilhomme}\footnotemark[1] \\
    \texttt{lgtclem@gmail.com} \\
    Aix-Marseille University
    \and
    \textbf{Andrew Lin}\footnotemark[1] \\
    \texttt{a\_lin@mit.edu} \\
    MIT
    \and
    \textbf{Florian Lorkowski}\footnotemark[1] \\
    \texttt{florian@lorkow.ski} \\
    University of Zurich
    \and
    \textbf{Marina Pérez del Valle} \footnotemark[1] \\
    \texttt{mperezdelval@umass.edu} \\
    University of Massachusetts Amherst
    \and
    \textbf{Radman Rakhshandehroo}\footnotemark[1] \\
    \texttt{rdmnr@student.ubc.ca} \\
    University of British Columbia
    \and
    \textbf{Patric Rommel}\footnotemark[1] \\
    \texttt{patric.rommel@itp1.uni-stuttgart.de} \\
    University of Stuttgart
    \and
    \textbf{Emanuel Ruzak}\footnotemark[1] \\
    \texttt{eruzak@dc.uba.ar} \\
    University of Buenos Aires
    \and
    \textbf{Nathan Theng}\footnotemark[1] \\
    \texttt{theng\_nathan@mail.fresnostate.edu} \\
    California State University, Fresno
    \and
    \textbf{Paul Yushin Rapoport}\footnotemark[2] \\
    \texttt{pyrapoport@uchicago.edu}\\
    University of Chicago
}
\date{May 6, 2026}
\sparround{Spring 2026}

\pgfplotsset{compat=1.18}
\begin{document}

\maketitle
\small \textit{Accepted to the Finding the Frame Workshop at RLC 2026}

\footnotetext[1]{All authors contributed equally and should be considered co-first authors; order is alphabetical, apart from Paul Rapoport, who was the mentor for the stream.}
\footnotetext[2]{Corresponding author: pyrapoport@uchicago.edu. All funding and logistical support was provided by Kairos and the SPAR program.}

\begin{abstract}
Classical reinforcement learning assumes the agent interacts with a fixed environment whose behavior does not depend on the agent's policy. This assumption breaks down in non-realizable settings where other actors might anticipate the agent's behavior -- most notably those environments crucial to AI safety, where any given agent interacts with predictors, human and other AI agents, and institutions. In such environments, the agent's model class fails to capture the world in which it operates. Under such misspecification, Classical Bayesian methods can produce confidently wrong posteriors, unreliable decisions, and unbounded regret, as realizability fails to obtain. Infra-Bayesianism is a decision-theoretic framework that addresses these failures by distinguishing ordinary probabilistic uncertainty -- where priors can be reasonably chosen -- from Knightian uncertainty, where no grounds exist for the construction of such a prior. Infra-Bayesianism does so by evaluating actions on their worst-case outcomes in environments, rather than from posterior expectations or weighted averaging.

We present the first proof-of-concept implementation of an infra-Bayesian reinforcement learning architecture for finite-outcome stateless decision problems. Our agent maintains a set of imprecise hypotheses, updates them using infra-Bayesian conditioning, and selects actions by maximizing worst-case expected value. We apply this implementation of the infra-Bayesian maximin decision process to an environment with Knightian uncertainty, and demonstrate a lower worst-case regret as compared to classical reinforcement learning agents. We also investigate Newcomb's problem and show that the infra-Bayesian agent picks the optimal strategy, outperforming classical decision theory agents. Our results provide a step towards reinforcement learning agents that remain robust under model misspecification and policy-dependent uncertainty.\end{abstract}

\section{Introduction}

Reinforcement learning (RL) is usually analyzed under the assumption that the agent interacts with a fixed environment: a Markov decision process (MDP), a partially observable Markov decision process (POMDP), or a Bayesian posterior over such models. This assumption is mathematically convenient and has enabled a large body of convergence and regret guarantees. However, it is poorly suited to settings in which the agent is embedded in an environment too complex to be represented by the agent, or environments containing other actors, that model the agent's behavior and thereby respond to the agent's policy rather than merely to its realized actions. Examples include autonomous vehicles whose driving style is anticipated by human drivers, recommendation systems whose users adapt to the system's known behavior, and decision-theoretic problems such as Newcomb's problem. In these settings, the environment is not merely unknown; it can be policy-dependent and non-realizable from the agent's point of view \citep{bell2021newcomblike}. 

Classical Bayesian RL provides a principled treatment of ordinary uncertainty by placing a prior over possible environments. Its strongest guarantees, however, rely on a \emph{realizability} or \emph{grain of truth} assumption: the true environment, or a sufficiently exact model of it, must lie in the agent's hypothesis class. This assumption is implausible for open-ended deployment. The real world contains the agent itself and other systems of comparable or greater complexity, so no tractable hypothesis class can be expected to contain a complete description of the environment. Under such misspecification, Bayesian agents can become confidently wrong, and value-based RL agents can fail to converge to optimal policies \citep{gelb2025inadequate_bayes}. 

Infra-Bayesianism (IB) was developed to address this problem by replacing precise Bayesian hypotheses with imprecise, worst-case-evaluated hypotheses. Instead of representing uncertainty only by a probability distribution over worlds, an IB agent may represent \emph{Knightian uncertainty} over possible worlds and evaluate actions by a lower expectation: the value guaranteed against the worst admissible member of the hypothesis class. It then selects the action that maximizes this worst-case expected value. This maximin structure is safety-relevant because it turns the agent's objective into a lower-bound guarantee, rather than an average-case prediction. IB also supplies an update rule for these objects that is designed to preserve dynamic consistency across sequential observations \citep{diffractor_introduction_2020,kosoy2020basic_inframeasure}. 

More broadly, IB belongs to the agent foundations and AI safety research agenda, which seeks mathematically precise models of embedded agency. From this perspective, non-realizability is not an edge case but a central feature of advanced agents operating in the real world: the agent is itself part of the environment it is trying to model.

Despite substantial theoretical work on IB and its role in the Learning-Theoretic Agenda for AI alignment, comparatively little work has translated its mathematical objects into concrete reinforcement learning agents. This leaves a gap between the formal promise of IB -- robust reasoning under non-realizability, Knightian uncertainty, and policy-dependent environments -- and the practical question of how an agent could actually represent, update, and plan within it. We present a proof-of-concept IB RL architecture. The implementation-level object is not a posterior distribution over MDPs, but a finite set of extremal affine evaluators whose lower envelope defines the agent's value estimate. This representation allows the agent to perform tractable lower-expectation evaluation, distinguish classical probabilistic mixtures from Knightian uncertainty, and apply dynamically consistent IB-style updates without enumerating full history trees.

Our contributions are:
\begin{enumerate}
    \item We formulate a finite-outcome infra-Bayesian reinforcement learning (IBRL) architecture based on a-measures and infradistributions, including classical and Knightian mixtures, lower-expectation evaluation, and IB-style updates.
    \item We show that the architecture recovers ordinary Bayesian behavior in the degenerate case where each infradistribution has a single minimal point and all uncertainty is classical.
    \item We demonstrate empirically that an IB agent outperforms a Bayesian agent in terms of worst-case performance in a Knightian uncertain environment.
    \item We show that IB decision theory obtains optimal rewards in simple policy-dependent environments, where classical decision theories fail to comprehensively model the causal structure.
\end{enumerate}

\section{Background}

\subsection{Classical and Bayesian reinforcement learning}

In RL, an agent repeatedly chooses actions and receives observations from an environment. The usual formalism is a Markov decision process,
which models the environment as a set of states through which the agent can transition by choosing from a set of actions, receiving a reward upon each transition. The agent aims to optimize its policy, to maximize the cumulative expected reward over future actions. Maximizing rewards is equivalent to minimizing \emph{regret}, the difference between the expected reward obtainable by the optimal policy and the actually realized reward.

Classical Bayesian RL places a prior over environments and updates by Bayes' rule, which is appropriate in realizable settings where the agent's hypothesis class contains the truth, and is the basis for many standard convergence arguments \citep{sutton2018reinforcement,ghavamzadeh2015bayesian}.
This assumption is a poor fit for embedded or open-world agents, where no computationally feasible hypothesis class can contain the exact environment. In such non-realizable settings, Bayesian updating still produces a posterior, but the posterior need not converge to a useful model for decision-making. This is one of the core motivations for IB in the Learning-Theoretic Agenda \citep{gelb2025inadequate_bayes,kosoy_learning-theoretic_2023}.

\subsection{Non-realizability and policy-dependent environments}

The central motivation for this work is that difficulties in RL arise not merely from statistical uncertainty. They also include model misspecification, Knightian uncertainty, and environments whose behavior depends on the agent's policy rather than only on its realized actions.

Policy-dependence is especially important for embedded agents. In ordinary RL, the environment is usually modeled as responding to the current state and action. In a policy-dependent environment, however, rewards or transitions may depend directly on the agent's policy. This can occur when other agents, predictors, markets, users, or institutions form expectations about the agent's behavior and respond to those expectations.

Newcomb-like decision processes formalize these environments by allowing rewards or transitions to depend on the policy itself. Bell et al.\ show that value-based RL agents in such environments can converge only to ratifiable policies, and may fail to converge to optimal policies at all \citep{bell2021newcomblike}. This suggests that policy-dependent environments require a decision-theoretic treatment that goes beyond ordinary value learning in fixed MDPs.

\subsection{Brief description of Newcomb's Problem}
Newcomb's problem (named after physicist William Newcomb) is a classic decision theory dilemma \citep{nozick1969newcomb}.  
In this problem, the agent is presented with a transparent box and an opaque box, and may choose either to take only the opaque box (one-boxing) or to take both boxes (two-boxing). The agent can see that the transparent box contains one thousand dollars. The opaque box contains either nothing or one million dollars, depending on a prediction that has already been made. Crucially, the prediction was about the agent's choice: the opaque box contains a million dollars if and only if the agent is predicted to take just the opaque box. The prediction is highly accurate, but not necessarily perfect. All aspects of the scenario are known to the agent. This scenario poses a dilemma because in it, the principle of dominance conflicts with the principle of expected-utility maximization. In particular, agents following causal or evidential decision theories come to different conclusions, and both fail to capture the causal structure of the environment accurately.

\subsection{Infra-Bayesianism at a high level}

Infra-Bayesianism generalizes Bayesian reasoning by allowing hypotheses to contain Knightian uncertainty. Instead of representing uncertainty only by a probability distribution over possible worlds, an IB agent may represent a set of admissible evaluators. The agent then evaluates a policy by its lower expectation: the value guaranteed under the least favorable admissible evaluator.

This distinction is important. A Bayesian agent averages over hypotheses using posterior probabilities. An IB agent can also represent ambiguity that should not be averaged away. In this sense, IB separates ordinary probabilistic uncertainty from Knightian uncertainty. The resulting decision rule is maximin: choose the policy whose worst admissible value is best.

This lower-bound perspective is especially relevant in safety-motivated settings. When the agent's model class may be misspecified, or when the environment may respond to the agent's policy, average-case posterior value may be the wrong object to optimize. IB instead provides a formalism for reasoning with partial information and worst-case guarantees.

\subsection{Affine measures and infradistributions}

The basic implementation-level object used in our implementation is an affine measure, or a-measure. In the finite non-signed setting considered here, an a-measure is a pair
\begin{equation}
a=(\lambda \mu,b),
\end{equation}
where $\mu$ is a probability measure over possible observation histories, $\lambda \geq 0$ is a scale factor, and $b \geq 0$ is an offset. Given a bounded return function $f$ over histories, the a-measure evaluates $f$ by
\begin{equation}
a(f)=\lambda \mathbb{E}_{\mu}[f]+b.
\end{equation}

The probability component $\mu$ represents ordinary stochastic uncertainty over histories. The scale $\lambda$ determines the weight assigned to that component. The offset $b$ records value associated with branches of the history tree that have been ruled out by observation. Intuitively, when an observation eliminates some branches, IB does not simply delete the value associated with those branches. Instead, their contribution is carried forward into the affine offset, which helps preserve dynamic consistency between ex ante and ex post evaluation.

An infradistribution $\Psi$ can be viewed, for our finite purposes, as a set of affine evaluators. Its lower expectation is
\begin{equation}
\underline{\mathbb{E}}_{\Psi}[f]
=
\inf_{a \in \Psi} a(f).
\label{eq:infraEV}
\end{equation}
Thus, a policy is evaluated by the worst admissible a-measure in the infradistribution. In reward-maximization language, the agent chooses a policy maximizing this lower expectation.

\subsection{Classical mixtures and Knightian uncertainty}

IB distinguishes between classical probabilistic uncertainty and Knightian uncertainty. Classical uncertainty is represented by a mixture. If $\Psi_i$ are infradistributions and $w_i$ are prior weights, their classical mixture is
\begin{equation}
\sum_i w_i \Psi_i
=
\Big\{
\sum_i w_i a_i \;:\; a_i \in \Psi_i
\Big\},
\qquad
\sum_i w_i = 1.
\label{eq:mixture}
\end{equation}
This corresponds to uncertainty that can be averaged over using prior weights, as in Bayesian reasoning.

Knightian uncertainty is different. It represents ambiguity that remains exposed to the outer infimum rather than being averaged away. A singleton infradistribution recovers ordinary probabilistic evaluation. A non-singleton infradistribution represents ambiguity over several admissible evaluators, where the value of a policy is determined by the least favorable one.

This distinction is central to the behavior of an IB agent. A Bayesian agent facing several possible models assigns probabilities to them and optimizes posterior expected value. An IB agent may instead treat several models as admissible without assigning meaningful probabilities between them, and then choose the policy whose lower expectation is highest.

\subsection{IB-style updating}

The IB update rule can be understood as a dynamically consistent analogue of Bayesian conditioning. After observing an event $L$, each a-measure is restricted to the observed branch. However, unlike ordinary Bayesian conditioning, the value associated with the unobserved branch is not simply discarded. Instead, it is transferred into the offset term.

To formalize this procedure, we must define the return function $g$. Unlike classical RL, where reward is typically a stepwise scalar $R(s, a)$ evaluated at each transition, IB evaluates policies using a bounded return (or loss) function $g$ defined over entire histories, formally, over infinite sequences of actions and observations (destinies). During evaluation, we will set $g \equiv f$. Let $\mu(g)$ denote the expected value of $g$ under the measure $\mu$. 
Viewing the observation $L$ as an indicator function for the realized event, the raw update takes the following schematic form:
\begin{equation}
(\lambda \mu,b)
\mapsto
(\lambda \mu L,\; b+\lambda \mu((1-L)g)).
\label{eq:RawIBUpdate}
\end{equation}
Here, $\mu L$ denotes the restriction of $\mu$ to the observed event. The complementary indicator $(1-L)$ isolates the branches of the history tree that were ruled out by the observation. The term $\lambda \mu((1-L)g)$ calculates the exact expected return of those unrealized branches, which gets added to the offset $b$. Intuitively, the offset records value that has already been settled by the observation, so that future lower-expectation evaluation remains dynamically consistent with the original ex ante evaluation.

After this restriction step, the resulting infradistribution is renormalized, such that
\begin{equation}
\underline{\mathbb{E}}_{\Psi}(0)=0,
\qquad
\underline{\mathbb{E}}_{\Psi}(1)=1.
\end{equation}
This normalization plays a role analogous to posterior normalization in Bayesian conditioning. In the special case where every infradistribution has exactly one minimal point and all uncertainty is represented by classical mixtures, the IB update reduces to ordinary Bayesian updating. This recovery of Bayes' rule is an important sanity check for both the formalism and our implementation.

An important property of the raw IB update~\eqref{eq:RawIBUpdate} is its linearity, implying that the IB update is a transformation in the space of a-measures mapping straight lines to straight lines. Intuitively, this means that the IB update does not produce new vertices. To reconstruct all relevant information contained in the infradistribution, the implementation thus only needs to track and update the extremal minimal points, which are the vertices of the set of minimal points.

\section{Related Work}

Existing IB work is mostly theoretical, covering inframeasures, update rules, dynamic consistency, learnability, and regret-style guarantees \citep{diffractor_introduction_2020,kosoy2020basic_inframeasure,gelb2025credal_sets}. Our work complements this by giving a finite RL implementation that explicitly represents infradistributions, performs IB-style updates, and plans using lower expectations.

We situate this contribution relative to three nearby areas: robust RL, imprecise probability and credal sets, and policy-dependent RL.

\subsection{Robust reinforcement learning}

Robust RL studies agents that optimize performance under worst-case uncertainty over environment models. In robust MDPs, uncertainty is usually represented as a set of possible transition kernels or reward functions, and the agent chooses a policy that performs well against the least favorable model in that set \citep{iyengar2005robust,nilim2005robust}.

This literature is closely related to our work because both approaches replace average-case evaluation with worst-case evaluation. The main difference is the representation of uncertainty. Robust MDPs typically retain a fixed, policy-independent environment model with uncertainty over transition or reward parameters. Our approach instead represents uncertainty using infradistributions over affine evaluators and updates those evaluators using the IB update rule.

\subsection{Imprecise probability and credal sets}

Imprecise probability theory represents uncertainty using sets of probability measures rather than a single precise distribution. This includes credal sets, lower and upper probabilities, and lower previsions. Walley's framework of coherent lower previsions is one of the standard foundations for reasoning with imprecise probabilities, while Levi's work on credal probability helped establish sets of probability functions as representations of belief states \citep{walley1991statistical,levi1980enterprise}.

Credal sets are closed and convex sets of probability distributions. They provide a simple representation of Knightian uncertainty, where uncertainty cannot be represented by a single precise probability distribution. Credal-set methods are closely related to our work because they also evaluate choices using lower expectations over a set of admissible probability models. However, IB is not merely a direct application of credal sets: in the finite setting considered here, an a-measure contains both a probabilistic component and an offset term. 

Recent domain-theoretic work on imprecise probability further emphasizes credal sets as structured objects supporting refinement and convergence \citep{edalat2026domain}; our work shares this finite-representation perspective, but implements lower-expectation planning using IB affine evaluators rather than ordinary probability measures.

\subsection{Policy-dependent and Newcomb-like environments}

Policy-dependent environments challenge the standard RL assumption that the environment is fixed independently of the agent's policy. Bell et al.\ formalize this issue for RL through Newcomb-like decision processes, where rewards or transitions may depend directly on the policy. They show that value-based RL agents in such environments cannot converge to non-ratifiable policies and may fail to converge to optimal policies \citep{bell2021newcomblike}. This motivates decision procedures that evaluate policies more directly, rather than only learning action values in a fixed environment. Our work provides a finite IB implementation for studying lower-expectation planning under non-realizability and policy-dependent uncertainty.

\section{Methods}

We implement a finite-outcome IB agent for stateless bandit and Newcomb-like decision problems.
The code is available at \url{https://github.com/SPAR-S26-IBRL-Stream/Project-Newcomb}.

The belief state of the agent is stored as a single infradistribution and a world model. The world model describes the type of environment in which the agent operates. In particular, the representation of the probability measure $\mu$ of each a-measure depends on the world model.

\subsection{Representing infradistributions}

General infradistributions are infinite sets of (signed) affine measures.
Operationally, measures can be discarded from this set if they are never needed to determine the lower expectation of any relevant function through equation~\eqref{eq:infraEV}.
For any infradistribution, only its minimal points contribute to the expectation values \citep{kosoy2020basic_inframeasure}.
Minimal points that can be written as non-trivial convex combinations of other minimal points also cannot contribute.
This means that only the extremal minimal points need to be stored, analogous to representing a convex polytope by its vertices.
This is the key computational representation used in our implementation.

Infradistributions are represented by their extremal minimal points, which is a finite set of a-measures.
Three constructions cover every belief state we use:
\begin{enumerate}
    \item Singleton: a set containing a single a-measure encodes a hypothesis without uncertainty. These infradistributions correspond to knowing the true environment exactly.
    \item Classical (Bayesian) mixture: a weighted combination of infradistributions, according to equation~\eqref{eq:mixture}. These represent classical probabilistic uncertainty. If each component is a singleton, this mixture reduces to standard Bayesian averaging.
    \item Knightian mixture: the set-union of the constituent infradistributions. This mixture carries no weights. It is evaluated by the worst-case across components.
\end{enumerate}
The two mixture operations compose and nest freely, which allows representing the rich belief states on which an IB agent operates.

\subsection{World models}\label{sec:world-model}
Conceptually, measures are probability distributions over infinite histories. Making them computationally tractable requires compressed representations.
Each type of environment uses a separate world model that defines how measures are represented and how predictions are derived from them, how past observations are represented, and how weighted mixtures of a-measures are computed. We focus on two world models: Bernoulli bandits and Newcomb-like problems.

In a one-armed Bernoulli bandit, the history can be represented as a pair of integers $(N,R)$, where $N$ is the number of times the arm was pulled and $R$ is the number of times a reward was obtained. This representation uses the fact that each round is independent, such that the order of observations does not matter. A non-mixed measure is represented by a fixed reward probability $p$. Mixed measures are represented as a set of pairs $(c_i,p_i)$, where $p_i$ is the probability and $c_i$ the weight of the $i^\text{th}$ component of the mixture and $\sum_i c_i=1$. Given such a history and measure, the probability of a certain branch is computed as \makebox{$\sum_i c_i (1-p_i)^{N-R} p_i^R$}. Predictions for future outcomes can be computed similarly. Learning proceeds by updating the history $(N,R)$, implicitly recreating Bayes' rule.
For a $k$-armed Bernoulli bandit, histories are represented by $k$ pairs $(N_j,R_j)$ and measures by $k$ sets of pairs $(c_{ji},p_{ji})$. Expectation values are computed per-arm as for the one-armed bandit. This factorization is possible because arms are independent of each other, i.e.\ observing one arm does not give any information about the others.

We consider Newcomb-like environments with imperfect predictors. The world model itself contains the entire reward matrix and the predictor accuracy, i.e.\ the agent knows the entire structure of the environment. In this setting, there is nothing to be learned. Therefore, measures and histories do not have an internal state and are not updated upon observations.

\subsection{Decision and update procedure}

The aim of the agent is to select the optimal policy, which might be non-deterministic, based on its belief state. The agent achieves this by discretizing policy space and maintaining a set of candidate policies $\Pi$. At every step of the interaction, the agent will:
\begin{enumerate}
    \item Compute the expected value of each policy $\pi \in \Pi$ as \makebox{$\underline{\mathbb{E}}_{\Psi(\pi)}[f] = \inf_{a \in \Psi(\pi)} a(f)$}, where $\Psi(\pi)$ is the infradistribution and $f$ is the reward function. The infradistribution can depend on the policy explicitly, such as in Newcomb-like environments, or implicitly, by sampling an action from the policy. With $|\Psi|=1$ this operation reduces to an ordinary expected value. With $|\Psi|>1$ it picks the worst-case environment.
    \item Select the optimal policy \makebox{$\pi^* = \mathop{\mathrm{argmax}}_{\pi \in \Pi} \underline{\mathbb{E}}_{\Psi(\pi)}[f]$}, and sample an action from $\pi^*$. Both the policy and the action are passed to the environment.
    \item Update its belief state upon receiving an observation from the environment. This includes raw updates of a-measures according to equation~\eqref{eq:RawIBUpdate}, updates of the probability measures as specified by the world model, and the renormalization step. Because of linearity of the raw update, no new extremal points appear, so it suffices to update the existing set.
\end{enumerate}

\section{Results and Discussion}

In the following sections, we introduce experiments that validate our implementation of the IB agent. First, we show that an IB agent reduces to a Bayesian agent when initialized equivalently in a standard stochastic bandit setting. We then demonstrate proper behavior under Knightian uncertainty; next, we validate behavior in Newcomb's problem. Finally, we show how an IB agent can meaningfully learn under Knightian uncertainty.

\subsection{Empirical validation that IB can reproduce classical behavior}
\label{section:validate-classical}

To validate our implementation, we test whether our IB implementation reproduces ordinary Bayesian behavior when the agents are initialized with identical hypothesis sets. We consider standard
two-armed Bernoulli bandits and compare a classical discrete Bayesian agent against an IB agent with a single \(a\)-measure. This \(a\)-measure contains the same finite Bernoulli hypothesis grid used by the classical agent, so the pessimistic minimum over \(a\)-measures is vacuous and the IB posterior predictive should reduce to the classical Bayesian
posterior predictive.

\begin{figure}[!htbp]
    \centering
    \input{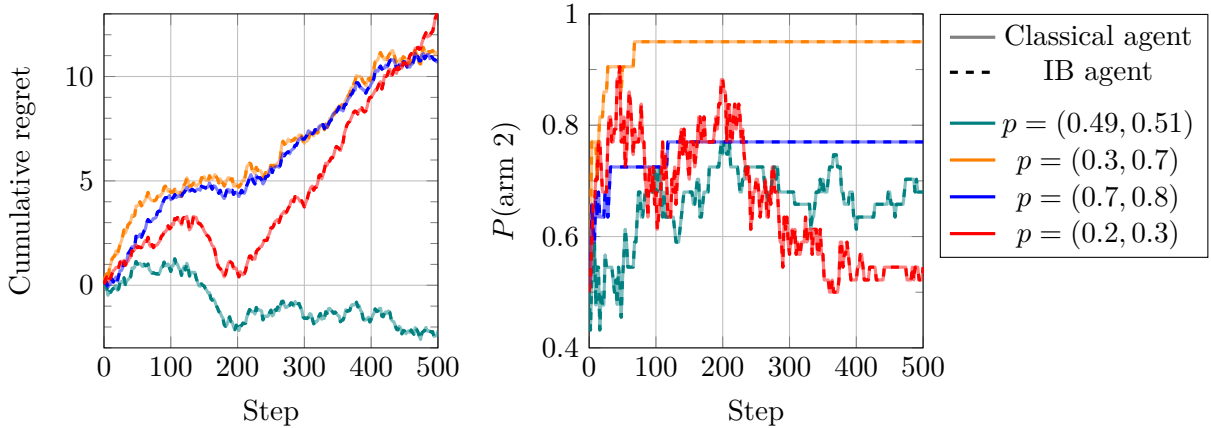}
    \caption{
    Cumulative regret (left) and probability of choosing the second arm (right) for the classical and IB agent in four different bandit environments. Solid, transparent curves correspond to the classical Bayesian agent and opaque, dashed curves to the IB agent. Colors indicate the different environments. }
    \label{fig:validate-classical}
\end{figure}

Figure~\ref{fig:validate-classical} shows the resulting cumulative regret and action probabilities across four stochastic, two-armed bandit environments. Both agents are initialized with a uniform prior over their hypotheses. In all four environments, the second arm is dominant, with the difference between the arm varying by environment.
All curves coincide exactly under matched random seeds, confirming that the IB agent recovers the classical Bayesian update and action rule in the single-measure setting.

\subsection{Knightian Uncertainty}\label{sec:ku-bandit}
To demonstrate Knightian uncertainty, we consider a two-armed, adversarial Bernoulli bandit environment. Each arm yields reward 1 with a probability chosen anew at the beginning of each step, and otherwise reward 0. The mechanism by which probabilities are chosen is unknown to the agent and may potentially be time-dependent or even adversarial. This makes it impossible to learn reward probabilities over multiple episodes. Past observations do not provide useful data for assigning a prior to the current interaction. 

\begin{figure}[h]
    \centering
    \pgfplotstableread{
  t classical  ib
  1  -0.20  -0.30
  2  -0.40  -0.60
  3   0.40   0.10
  4   1.20  -0.20
  5   2.00   0.50
  6   2.80   1.20
  7   3.60   0.90
  8   3.40   0.60
  9   4.20   0.30
 10   5.00   1.00
 11   5.80   1.70
 12   6.60   2.40
 13   7.40   3.10
 14   8.20   3.80
 15   9.00   4.50
 16   9.80   5.20
 17   9.60   4.90
 18  10.40   4.60
 19  11.20   5.30
 20  11.00   5.00
 21  11.80   5.70
 22  12.60   6.40
 23  13.40   7.10
 24  14.20   7.80
 25  15.00   8.50
 26  15.80   9.20
 27  15.60   8.90
 28  16.40   9.60
 29  17.20  10.30
 30  17.00  10.00
 31  17.80  10.70
 32  18.60  11.40
 33  19.40  11.10
 34  20.20  11.80
 35  21.00  11.50
 36  21.80  11.20
 37  22.60  11.90
 38  23.40  12.60
 39  24.20  13.30
 40  25.00  14.00
 41  24.80  13.70
 42  24.60  13.40
 43  25.40  14.10
 44  26.20  14.80
 45  26.00  14.50
 46  26.80  15.20
 47  27.60  15.90
 48  27.40  15.60
 49  27.20  15.30
 50  27.00  15.00
 51  26.80  14.70
 52  27.60  15.40
 53  28.40  16.10
 54  29.20  15.80
 55  30.00  16.50
 56  30.80  17.20
 57  31.60  17.90
 58  32.40  18.60
 59  33.20  19.30
 60  33.00  19.00
 61  33.80  19.70
 62  34.60  20.40
 63  35.40  21.10
 64  35.20  20.80
 65  36.00  21.50
 66  36.80  21.20
 67  37.60  20.90
 68  38.40  21.60
 69  39.20  22.30
 70  39.00  22.00
 71  38.80  21.70
 72  39.60  22.40
 73  39.40  22.10
 74  39.20  21.80
 75  40.00  22.50
 76  40.80  22.20
 77  41.60  22.90
 78  41.40  22.60
 79  41.20  22.30
 80  41.00  22.00
 81  41.80  22.70
 82  42.60  23.40
 83  43.40  24.10
 84  44.20  24.80
 85  45.00  25.50
 86  44.80  25.20
 87  45.60  24.90
 88  45.40  24.60
 89  45.20  24.30
 90  46.00  25.00
 91  45.80  24.70
 92  46.60  25.40
 93  47.40  26.10
 94  48.20  26.80
 95  49.00  26.50
 96  49.80  27.20
 97  50.60  27.90
 98  50.40  27.60
 99  51.20  28.30
100  52.00  29.00
}\data

\begin{tikzpicture}[
  baseline=(current bounding box.center),
  /pgfplots/every axis/.append style={
    width=6cm,
    height=6cm,
    xmajorgrids,
    ymajorgrids,
    x label style={at={(axis description cs:0.5,-0.125)},anchor=north},
    y label style={at={(axis description cs:-0.125,0.5)},anchor=south},
  }
]
    \begin{axis}[
        name=envplot,
        xmin=0,xmax=1,xtick distance=0.5,minor x tick num=4,
        ymin=0,ymax=1,ytick distance=0.5,minor y tick num=4,
        xlabel={Probability $p_1$},
        ylabel={Probability $p_2$},
    ]
        \fill[green,opacity=0.5] (axis cs:0,0) -- (axis cs:1,1) -- (axis cs:0,1) -- cycle;
        \fill[blue ,opacity=0.5] (axis cs:0,0) -- (axis cs:1,1) -- (axis cs:1,0) -- cycle;
        \path[draw=black,thick,fill=white,opacity=0.85]
            (axis cs:0.3,0.4) rectangle (axis cs:0.7,0.8);
        \draw[thick,dashed] (axis cs:0,0) -- (axis cs:1,1);
        \fill[red] (axis cs:0.3,0.4) circle[radius=0.055cm];

        \node[anchor=base] at (axis cs:0.28,0.85) {$p_2 > p_1$};
        \node[] at (axis cs:0.73,0.10) {$p_1 > p_2$};
    \end{axis}

    \begin{axis}[
        name=regretplot,
        at={(envplot.east)},
        anchor=west,
        xshift=1.5cm,
        xmin=0,xmax=100,xtick distance=50,minor x tick num=4,
        ymin=0,ymax=55,ytick distance=25,minor y tick num=4,
        xlabel={Episode},
        ylabel={Cumulative Regret},
        legend style={
            at={(rel axis cs:1.05,1)},
            anchor=north west
        }
    ]

        \fill[teal,opacity=0.5] (axis cs:0,0) -- (axis cs:100,50) -- (axis cs:100,0) -- cycle;
        \fill[orange,opacity=0.5] (axis cs:0,0) -- (axis cs:100,30) -- (axis cs:100,0) -- cycle;
    
        \addplot[teal,very thick,forget plot,line join=round] table[x=t,y=classical] \data;
        \addplot[orange,very thick,forget plot,line join=round] table[x=t,y=ib] \data;
    
        \addlegendimage{teal,opacity=0.5,line width=5pt}
        \addlegendentry{Classical allowed}
        \addlegendimage{teal,thick}
        \addlegendentry{Classical worst-case}
        \addlegendimage{orange,opacity=0.5,line width=5pt}
        \addlegendentry{IB allowed}
        \addlegendimage{orange,thick}
        \addlegendentry{IB worst-case}
    
    \end{axis}

\end{tikzpicture}
    \caption{Left: Visualization of environment space $(p_1,p_2)$. In the blue (green) area, arm 1 (2) yields the higher expected reward. The white box indicates the region that is allowed by the constraint. The red dot indicates the worst allowed environment. Right: cumulative regret of classical and IB agents. The shaded areas show the theoretically allowed ranges. The lines show simulated results from a single roll-out of the worst-case configuration for each agent.}
    \label{fig:ku}
\end{figure}
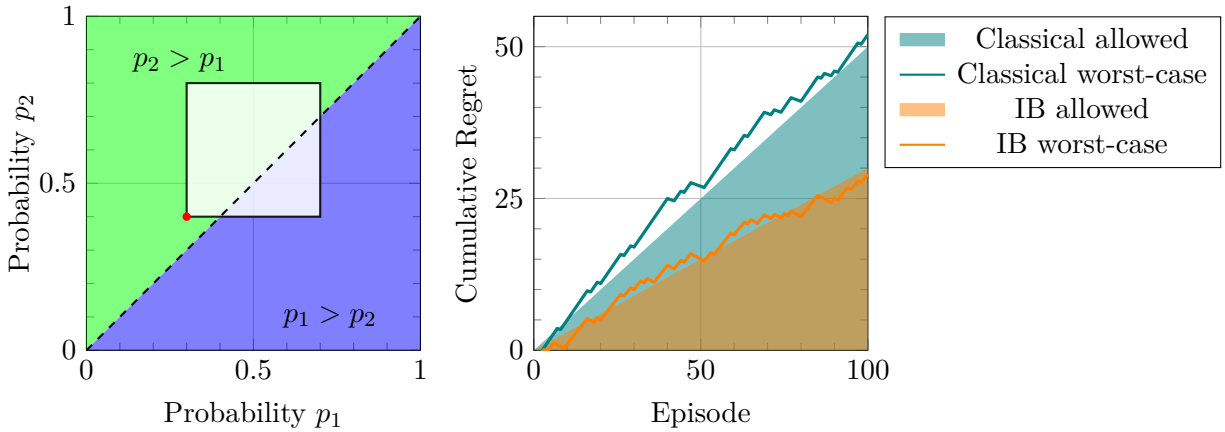

The reward probabilities are constrained to the range $p_1 \in [0.3, 0.7]$ and $p_2 \in [0.4, 0.8]$. The left of figure~\ref{fig:ku} displays the space of possible environments. There are two regions: in the upper triangle of the diagram $p_2 > p_1$, and thus it is optimal to pull arm 2. In the lower triangle, the opposite is true. For decision making, it only matters whether the agent believes it is in the upper or lower region. Both regions are possible, within the constraint.

A classical Bayesian agent cannot act from the interval constraints alone. It must first replace them with a precise prior over environments. Different choices of this prior can lead to different policies. For example, a prior concentrated on an environment where \(p_1 > p_2\) recommends arm 1, while a prior concentrated on an environment where \(p_2 > p_1\) recommends arm 2. In this experiment, we illustrate this prior-dependence by considering classical agents whose priors are point masses at the corners of the allowed set; of course, this should not be read as the only possible classical choice. A uniform prior over the allowed set would recommend arm 2 in this example, matching the IB action. The point is instead that the classical agent’s behavior depends on an additional prior-selection assumption that is not specified by the interval constraints themselves. By contrast, the IB agent can represent the constraint directly as Knightian uncertainty, which will lead it to maximize the worst-case outcome. The worst allowed environment is $(p_1,p_2) = (0.3,0.4)$, as indicated in the figure. In this environment $p_2 > p_1$, so the agent will always pull arm 2. It is guaranteed to achieve an average reward of $0.4$, and does not risk getting $0.3$ on arm 1.

The exact reward and regret realized depend on the true environment. The right panel of figure~\ref{fig:ku} shows the possible ranges of regret achieved by the two agents, for any true environment consistent with the constraint. Also shown are simulated regret curves for the worst-case configurations for each agent. We find that the worst-case outcome for the IB agent realizes a lower regret than the worst-case of the classical agent.

Note that the results here are not meant to demonstrate a meaningful form of learning, either by the IB or classical agents. Indeed, even if arm 1 repeatedly gives higher rewards, an IB agent will not update to exploit them, which is the correct behavior under Knightian uncertainty. Rewards could be controlled by an unknown or adversarial mechanism that makes arm 1 look attractive, only to give it a lower reward when the agent chooses it. Sticking to arm 2 is therefore the robust and safe strategy in this scenario. We present a stochastic bandit setting, in which the IB agent does learn despite Knightian uncertainty in section~\ref{section:trap-bandits}.

\subsection{Policy dependence}

Policy dependence is investigated via Newcomb's problem with an imperfect predictor. We consider a setting in which the transparent box always contains \$1. The opaque box contains \$10 if the agent is predicted to one-box, and \$0 otherwise. The resulting payoffs are shown in table~\ref{tab:newcomb}. The agent chooses a policy that one-boxes with probability $p$. A predictor with accuracy $\alpha \in [0.5,1]$ reads the agent's policy and predicts one-boxing with probability $p\cdot(2\alpha-1) + 0.5\cdot(2-2\alpha)$. A perfect predictor ($\alpha=1$) predicts one-boxing at the same rate $p$ that the agent one-boxes. A random predictor ($\alpha=0.5$) guesses at random, thus its prediction is independent of the agent’s policy.

\begin{table}[!htbp]
    \centering
    \begin{tabular}{l|c|c}
            & Predicted & Predicted \\
            & one-box & two-box \\ \hline
             One-box & 10 & 0  \\ \hline
             Two-box & 11 & 1
    \end{tabular}
    \caption{Reward matrix for Newcomb's problem }
    \label{tab:newcomb}
\end{table}

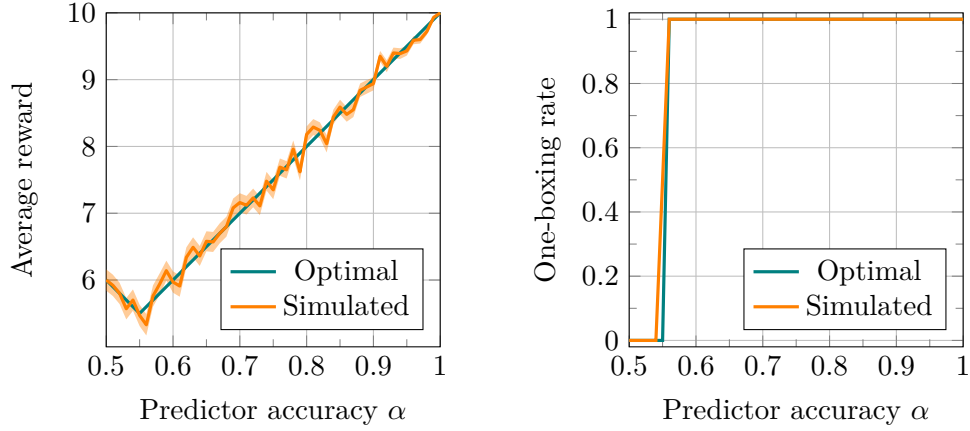
\begin{figure}[h]
    \centering
    \pgfplotstableread{
acc reward_opt reward_sim rate_sim rate_opt reward_sim_err
0.50  6.00  6.00 1.0 1.0 0.16
0.51  5.90  5.92 1.0 1.0 0.16
0.52  5.80  5.80 1.0 1.0 0.16
0.53  5.70  5.57 1.0 1.0 0.16
0.54  5.60  5.70 1.0 1.0 0.16
0.55  5.50  5.47 0.5 1.0 0.16
0.56  5.60  5.33 0.0 0.0 0.16
0.57  5.70  5.76 0.0 0.0 0.16
0.58  5.80  5.94 0.0 0.0 0.16
0.59  5.90  6.14 0.0 0.0 0.15
0.60  6.00  5.97 0.0 0.0 0.16
0.61  6.10  5.91 0.0 0.0 0.16
0.62  6.20  6.34 0.0 0.0 0.15
0.63  6.30  6.49 0.0 0.0 0.15
0.64  6.40  6.36 0.0 0.0 0.15
0.65  6.50  6.58 0.0 0.0 0.15
0.66  6.60  6.57 0.0 0.0 0.15
0.67  6.70  6.70 0.0 0.0 0.15
0.68  6.80  6.80 0.0 0.0 0.15
0.69  6.90  7.08 0.0 0.0 0.14
0.70  7.00  7.16 0.0 0.0 0.14
0.71  7.10  7.12 0.0 0.0 0.14
0.72  7.20  7.23 0.0 0.0 0.14
0.73  7.30  7.11 0.0 0.0 0.14
0.74  7.40  7.48 0.0 0.0 0.14
0.75  7.50  7.35 0.0 0.0 0.14
0.76  7.60  7.69 0.0 0.0 0.13
0.77  7.70  7.65 0.0 0.0 0.13
0.78  7.80  7.96 0.0 0.0 0.13
0.79  7.90  7.62 0.0 0.0 0.13
0.80  8.00  8.18 0.0 0.0 0.12
0.81  8.10  8.29 0.0 0.0 0.12
0.82  8.20  8.24 0.0 0.0 0.12
0.83  8.30  8.04 0.0 0.0 0.13
0.84  8.40  8.44 0.0 0.0 0.11
0.85  8.50  8.59 0.0 0.0 0.11
0.86  8.60  8.48 0.0 0.0 0.11
0.87  8.70  8.55 0.0 0.0 0.11
0.88  8.80  8.84 0.0 0.0 0.10
0.89  8.90  8.89 0.0 0.0 0.10
0.90  9.00  8.94 0.0 0.0 0.10
0.91  9.10  9.35 0.0 0.0 0.08
0.92  9.20  9.20 0.0 0.0 0.09
0.93  9.30  9.40 0.0 0.0 0.08
0.94  9.40  9.39 0.0 0.0 0.08
0.95  9.50  9.43 0.0 0.0 0.07
0.96  9.60  9.59 0.0 0.0 0.06
0.97  9.70  9.60 0.0 0.0 0.06
0.98  9.80  9.72 0.0 0.0 0.05
0.99  9.90  9.93 0.0 0.0 0.03
1.00 10.00 10.00 0.0 0.0 0.00
}\data

\begin{tikzpicture}[
  baseline=(current bounding box.center),
  /pgfplots/every axis/.append style={
    width=6cm,
    height=6cm,
    xmin=0.5,xmax=1,xtick distance=0.1,minor x tick num=1,
    xlabel={Predictor accuracy $\alpha$},
    xmajorgrids,
    ymajorgrids,
    x label style={at={(axis description cs:0.5,-0.125)},anchor=north},
    y label style={at={(axis description cs:-0.175,0.5)},anchor=south},
  }
]
    \begin{axis}[
        name=rewardplot,
        ymin=5,ymax=10,ytick distance=1,minor y tick num=0,
        ytickmin=5.5,
        ylabel=Average reward,
        legend style={at={(rel axis cs:0.95,0.05)},anchor=south east}
    ]

        \addplot[teal,very thick] table[x=acc,y=reward_opt] \data;
        \addlegendentry{Optimal}
        \addplot[name path=upp,draw=none,forget plot]
            table[x=acc,y expr={\thisrow{reward_sim}+\thisrow{reward_sim_err}}] \data;
        \addplot[name path=low,draw=none,forget plot]
            table[x=acc,y expr={\thisrow{reward_sim}-\thisrow{reward_sim_err}}] \data;
        \addplot[orange,opacity=0.4,forget plot] fill between [of=low and upp];
        \addplot[orange,very thick,line join=round] table[x=acc,y=reward_sim] \data;
        \addlegendentry{Simulated}
    \end{axis}

    \begin{axis}[
        name=rateplot,
        at={(rewardplot.east)},
        anchor=west,
        xshift=2.5cm,
        ymin=-0.02,ymax=1.02,ytick distance=0.2,minor y tick num=1,
        ylabel={One-boxing rate},
        legend style={at={(rel axis cs:0.95,0.05)},anchor=south east}
    ]

    \addplot[teal,very thick] table[x=acc,y expr={1-\thisrow{rate_opt}}] \data;
    \addplot[orange,very thick] table[x=acc,y expr={1-\thisrow{rate_sim}}] \data;
    \legend{Optimal,Simulated}

    \end{axis}

\end{tikzpicture}
    \caption{Average reward (left) and one-boxing rate (right) in Newcomb's problem as a function of the predictor accuracy. Shown are both optimal and simulated values, averaged over 1000 episodes. For $\alpha=0.55$, the reward is independent of the one-boxing rate and thus every rate is optimal.}
    \label{fig:newcomb}
\end{figure}

We implement this setting using the Newcomb-like world model described in section~\ref{sec:world-model}. The optimal policy and reward are shown in figure~\ref{fig:newcomb}. For a sufficiently accurate predictor ($\alpha > 0.55$), one-boxing is the optimal strategy. For a nearly random predictor ($\alpha < 0.55$), two-boxing is optimal. The figure also shows simulated results from our implementation. We find that the agent consistently selects the optimal policy and achieves the corresponding optimal reward.

An agent following causal decision theory would two-box in Newcomb's problem, arguing that its decision cannot influence the already fixed box contents after the predictor's move, thus missing out on the large reward when the predictor is sufficiently accurate. An agent following evidential decision theory also one-boxes in this scenario, but for a different reason. It fails to accurately model the causal structure of the problem and can misbehave in similar scenarios.  This scenario demonstrates IB decision theory selecting the optimal action, and for the correct reason, given full information of the reward structure. The same approach also achieves optimal behavior in the (Asymmetric) Death in Damascus and Coordination Game scenarios, some of which require mixed strategies \citep{bell2021newcomblike}.

\subsection{Trap Bandit Experiments}
\label{section:trap-bandits}

We consider a simple experiment illustrating how a robust IB learner can be useful even in a stateless stochastic bandit setting. The environment has two arms, indexed by \(i \in \{1,2\}\). At the start of each run, the arm reward probabilities \((p_1,p_2)\) are sampled uniformly from
\[
    \{(0.3,0.7), (0.7,0.3)\},
\]
so that one arm has significantly higher expected reward than the other. Independently, the world is sampled to be either safe or risky, with probability \(\alpha\) of being risky and probability \(1-\alpha\) of being safe.

In a safe world, each arm is Bernoulli with reward probability \(p_i\). In a risky world, the arm with the larger reward probability is also the trapped arm. Pulling this arm yields a catastrophic reward of \(-1000\) with probability \(p_{\mathrm{cat}}\), yields reward \(1\) with probability \(p_i\), and yields reward \(0\) otherwise. The lower-probability arm remains Bernoulli. Thus each run is generated as follows:
\[
\begin{array}{ll}
\text{sample} & (p_1,p_2) \sim \mathrm{Unif}\{(0.3,0.7),(0.7,0.3)\},\\
\text{sample} & \text{world type} \sim \mathrm{Bernoulli}(\alpha),\\[2mm]
\text{if safe:} & r_i \sim \mathrm{Bernoulli}(p_i),\\[1mm]
\text{if risky:} &
\begin{cases}
\text{the trap arm } \arg\max_i p_i \text{ yields } -1000 \text{ with probability } p_{\mathrm{cat}},\\
\text{the trap arm yields } 1 \text{ with probability } p_i,\\
\text{the trap arm yields } 0 \text{ otherwise},\\
\text{the other arm yields } r_i \sim \mathrm{Bernoulli}(p_i).
\end{cases}
\end{array}
\]

We compare a classical Bayesian agent with an infra-Bayesian agent using the same joint hypothesis machinery. Bayesian agents represent uncertainty using classical prior distributions. The infra-Bayesian agent instead maintains Knightian uncertainty over the safe and risky world families, while retaining ordinary Bayesian uncertainty via prior probabilities over \((p_1,p_2)\) within each family.

The experiments vary the relationship between the true risky-world probability, \(\alpha_{\mathrm{DGP}}\), and the Bayesian agent's point prior, \(\alpha_{\mathrm{prior}}\). In the mostly risky setting, we set \(\alpha_{\mathrm{DGP}}=0.99\). We first consider a correctly specified Bayesian prior, \(\alpha_{\mathrm{prior}}=0.99\), and then a severely misspecified prior, \(\alpha_{\mathrm{prior}}=0.01\). Across these conditions, the infra-Bayesian agent uses the same classical prior over \((p_1,p_2)\) as the Bayesian agents, but maintains Knightian uncertainty over whether the world is safe or risky.

\begin{figure}[htb]
    \centering
    \includegraphics[width=\textwidth]{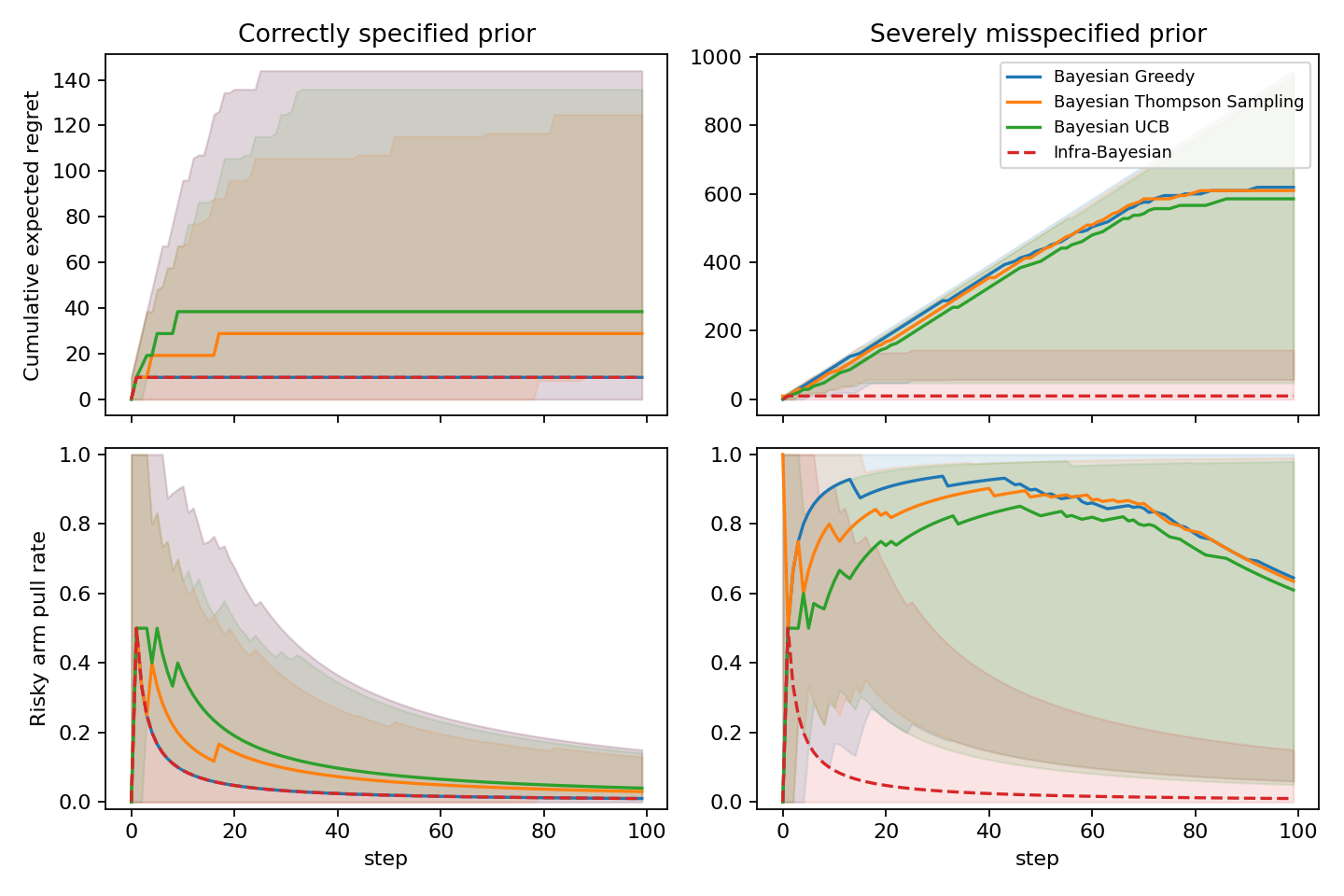}
    \caption{Comparing the performance of infra-Bayesian and classical Bayesian agents (with either greedy or Thompson Sampling exploration strategies) in the trap bandit setting. The first column shows results for a correctly specified Bayes prior condition; the second for a severely misspecified Bayes prior condition. The first row shows cumulative expected regret, and the second shows the average pull rate of the risky (trap) arm.}
    \label{fig:trapped-bandits}
\end{figure}

For Bayesian agents, we evaluate two exploration strategies: greedy action selection and Thompson sampling. The infra-Bayesian agent uses greedy action selection with respect to its robust lower values, with uniform tie-breaking. Regret is measured relative to the best policy with full knowledge of the true world. We report cumulative expected regret percentiles and trapped-arm pull-rate percentiles. Results are shown in Figure \ref{fig:trapped-bandits} and Table \ref{tab:trap-bandit-results}. In Figure \ref{fig:trapped-bandits}, solid lines show medians and shaded bands show the empirical 5th--95th percentile range across sampled risky-world runs at each time step; in Table \ref{tab:trap-bandit-results}, brackets instead give bootstrap confidence intervals for the final-time cumulative-regret percentiles, obtained by resampling worlds. 

When the Bayesian prior is correctly specified in a mostly risky worlds setting, the greedy Bayesian and infra-Bayesian agents behave nearly identically. This is expected: when the risky-world probability is high, expected-value maximization already favors the conservative action. The infra-Bayesian agent nevertheless obtains this behavior without committing to a point prior over the safe/risky model class. Under severe misspecification, however (column 2), the Bayesian agents initially treat the world as mostly safe, repeatedly pulling the high-reward trapped arm, and incurring much larger expected regret and high catastrophe rates. Thompson sampling slightly reduces regret in this misspecified condition, but does not remove the failure mode.

These results show infra-Bayesian agents matching or outperforming classical Bayesian agents in risky worlds, which raises a natural question: at what cost? Table~\ref{tab:trap-bandit-results} includes an alternative, mostly-safe scenario with \(\alpha_{\mathrm{DGP}}=0.01\). In this setting, the infra-Bayesian agent incurs substantially higher regret than classical agents, reflecting the cost of maintaining Knightian uncertainty over the risky-world hypothesis in a relatively safe setting.

\begin{table}[t]
\centering
\small
\begin{tabular}{lll rcc}
\toprule
$\alpha_{\mathrm{DGP}}$ & $\alpha_{\mathrm{prior}}$ & Agent & Catastrophe rate & p50, 95\% CI & p95, 95\% CI \\
\midrule
0.99 & n/a & infra\_bayesian & 0.040 & 9.60 [9.60, 9.60] & 144.00 [96.48, 183.36] \\
0.99 & 0.99 & bayes\_greedy & 0.040 & 9.60 [9.60, 9.60] & 144.00 [96.48, 183.36] \\
0.99 & 0.01 & bayes\_greedy & 0.650 & 609.60 [508.80, 739.20] & 960.00 [950.40, 960.00] \\
0.99 & 0.99 & bayes\_thompson & 0.075 & 28.80 [28.80, 38.40] & 124.80 [96.00, 154.56] \\
0.99 & 0.01 & bayes\_thompson & 0.645 & 595.20 [499.20, 739.20] & 950.40 [940.80, 950.40] \\
0.01 & n/a & infra\_bayesian & 0.000 & 39.60 [39.60, 39.60] & 40.00 [40.00, 40.00] \\
0.01 & 0.01 & bayes\_greedy & 0.015 & 0.40 [0.40, 0.40] & 4.80 [2.84, 6.40] \\
0.01 & 0.01 & bayes\_thompson & 0.015 & 1.60 [1.20, 1.60] & 6.82 [4.42, 7.60] \\
\bottomrule
\end{tabular}
\caption{Final cumulative expected-regret percentiles with bootstrap confidence intervals.}
\label{tab:trap-bandit-results}
\end{table}

\section{Conclusion, Limitations, and Future Work}

We present a proof-of-concept IB reinforcement learning architecture for finite-outcome stateless decision problems. Our design implements a-measures, infradistributions, classical and Knightian mixtures, and the IB conditioning rule, while remaining computationally tractable by storing only extremal minimal points and using optimized representations for histories and belief states. We find that the architecture recovers standard Bayesian behavior as a special case, confirming that IB generalizes rather than replaces classical reasoning. In a Knightian-uncertain bandits setting, the IB agent identifies and optimizes against the worst-case environment within the admissible set, while classical agents depend on arbitrary priors. This demonstrates concretely that the distinction between probabilistic and Knightian uncertainty, central to the IB formalism, produces meaningfully different agent behavior in settings where classical methods may be ambiguous. More broadly, real-world agents must operate under misspecification and policy-dependent environments. In such settings, optimizing expected value can lead to brittle behavior, while worst-case optimization offers robustness. Bridging IB theory to concrete agent implementations is a step toward RL systems that are robust by design.

Our implementation is restricted to finite outcomes, nonnegative a-measures, and small hypothesis spaces. Scaling to continuous state spaces, large hypothesis classes, and function approximation remains open. Future work -- some of which is already underway -- would seek to address these concerns, most notably permitting optimization of multi-step decision processes under Knightian uncertainty and fully leveraging IB's advantages in keeping track of multiple possible environments. Nevertheless, the results show that IB reasoning can be translated into a working RL agent. Additionally, our regret bounds, while an improvement over those of classical RL, remain linear, though those regret bounds for classical RL presuppose that a classical agent can converge to a ratifiable policy at all.

\clearpage

\bibliographystyle{plainnat}
\bibliography{references}

\clearpage
\end{document}